\documentclass[letterpaper, 10 pt, conference]{ieeeconf}  

\IEEEoverridecommandlockouts                              

\usepackage{graphicx} 
\usepackage{amsmath} 
\usepackage{amssymb}  
\usepackage{hyperref}
\usepackage{booktabs}
\usepackage{xcolor} 

\title{\LARGE \bf
A Framework for Active Haptic Guidance Using Robotic Haptic Proxies
}

\author{Niall L. Williams$^{\dagger 1}$, Nicholas Rewkowski$^{\dagger 2}$, Jiasheng Li$^{1}$, and Ming C. Lin$^{1}$
\thanks{*This work is partially supported by National Science Foundation and Lin's Barry Mersky and Capital One E-Nnovate Endowed Professorship.}
\thanks{$^{1}$Authors are with the Department of Computer Science, University of Maryland, College Park.
        {\tt\small \{niallw, jsli, lin\}@umd.edu}
}%
\thanks{$^{2}$Email: {\tt\small \{nick.rewkowski\}@gmail.com}}%
\thanks{
$^\dagger$Equal contribution
}%
\\
\href{https://gamma.umd.edu/active_haptic_guidance/}{\textcolor{blue}{\texttt{https://gamma.umd.edu/active\_haptic\_guidance/}}}
}

\begin{document}

\maketitle

\thispagestyle{empty}
\pagestyle{empty}

\begin{abstract}

Haptic feedback is an important component of creating an immersive mixed reality experience.
Traditionally, haptic forces are rendered in response to the user's interactions with the virtual environment.
In this work, we explore the idea of rendering haptic forces in a \textit{proactive} manner, with the explicit intention to influence the user's behavior through compelling haptic forces.
To this end, we present a framework for \textit{active haptic guidance} in mixed reality, using one or more robotic haptic proxies to influence user behavior and deliver a safer and more immersive virtual experience.
We provide details on common challenges that need to be overcome when implementing active haptic guidance, and discuss example applications that show how active haptic guidance can be used to influence the user's behavior.
Finally, we apply active haptic guidance to a virtual reality navigation problem, and conduct a user study that demonstrates how active haptic guidance creates a safer and more immersive experience for users.

\end{abstract}



\section{INTRODUCTION}

In mixed reality (MR), the user is at least partially immersed in a 3D, computer-generated environment.
Included within the mixed reality spectrum are augmented reality and virtual reality (VR).
A major factor that makes MR a unique medium is that it is interactive\textemdash the user is able to interact with the virtual environment (VE) through position-tracking sensors that update the VE according to the user's movements in the physical environment (PE).
For example, when the user moves their head in the real world, the position of the camera in the virtual world moves as well.
Interactions like these help to make users feel like they are really in the VE that they see through the head-mounted display (HMD).
One key component to increasing the user's sense of presence in a VE is to improve the perceptual stimuli matching \cite{hendrix1996presence}, wherein the user is provided with perceptual information that matches their actions (e.g. the viewing perspective updates as the user's head moves).
In this work, we focus on the sense of touch provided by \textcolor{black}{kinesthetic} haptic feedback and how we can use robots to provide more realistic haptic sensations to improve the sense of immersion and safety in MR.

\begin{figure}[!ht]
  \centering
  \includegraphics[width=0.95\linewidth]{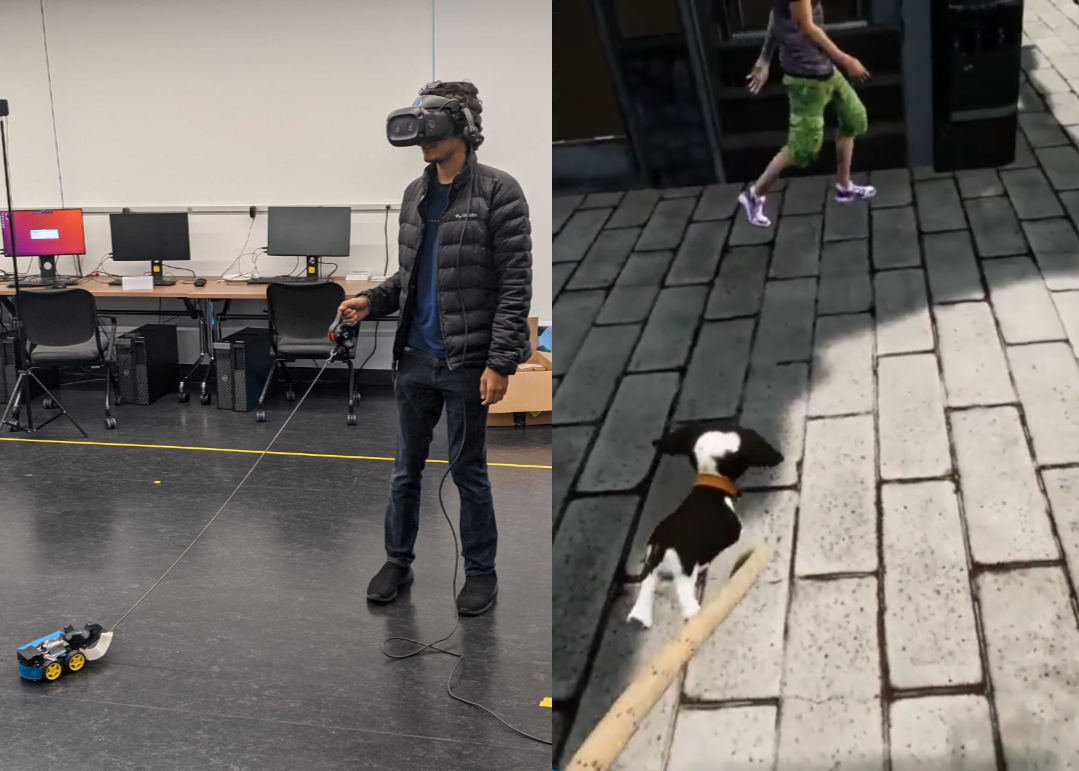}
  \caption{An image of a user in the physical environment (left) and virtual environment (right) in our implementation of active haptic guidance. The user is tethered to a robot in the physical environment and to a virtual dog companion in the virtual environment. The robot provides haptic feedback to the user according to the virtual companion's movements, which improves the user's sense of presence in the virtual world and encourages the user to avoid the boundaries of the virtual reality system's tracked space.}
  \label{fig:teaser}
  \vspace*{-1em}
\end{figure}

Robotic technology has in fact been used to provide haptic feedback in MR
to improve the sense of virtual touch and virtual manipulation \cite{hoenig2015mixed}.
For example, MR can enhance robotics via telepresence, wherein humans can remotely operate robot to high precision using immersive controls afforded by VR.

In this paper, we introduce the possibility of using robots to enhance the virtual experience through haptic feedback.
Specifically, we use robots to \textit{guide} the user as they navigate through a VE, and reconfigure and \textit{virtually expand} the PE to align with the VE; we achieve this through manual haptic feedback that directs the user's locomotion behavior in the VE, thereby making the virtual experience more immersive and safer.
To this end, we introduce the concept of \textit{active haptic guidance}, which describes the problem of reconfiguring one or multiple robots in the PE in real time such that they provide haptic feedback to guide the user and influence their actions and motion in the VE, with the ultimate goal of improving the user's safety or level of immersion in MR.
\textcolor{black}{Unlike traditional haptic guidance \cite{feygin2002haptic} that requires users to follow the haptic forces with the goal of teaching them specific skills, our framework maintains the user's freedom to explore and interact with the VE however they wish, with the goal of enhancing their virtual experience.}
One major challenge with robots for active haptic feedback in MR is that the physical robots and their virtual counterparts must be co-located \textit{relative to the user}, in order to provide the correct haptic feedback that aligns with the virtual counterpart.
This problem can be exacerbated when the environments/interactions are dynamic (i.e. the physical and virtual haptic proxy must move synchronously) or when there is a decoupling between the user's physical and virtual locations (as is common with some VR interaction techniques like redirected walking~\cite{razzaque2001redirected}).

\textbf{Main contributions:} We introduce the concept of \textit{active haptic guidance} for  improved {\em virtual locomotion}, and conduct a user study to show an example of how active haptic guidance can be used to improve a user's safety and feelings of immersion in a virtual experience.
Our framework is general, so it can be applied to use cases other than locomotion.
Our main contributions include:
\begin{itemize}
    \item A formal description of the \textit{active haptic guidance} problem and details on common challenges that are faced when implementing active haptic guidance. Active haptic guidance involves using robots to \textcolor{black}{\textit{proactively}} provide realistic haptic feedback to users in mixed reality, with the goal of influencing users' behaviors to improve safety and/or sense of presence in the VE.
    
    \item An prototype realization and user study showing the benefits of active haptic guidance. In our study, participants completed a virtual navigation task using real walking, either with or without active haptic guidance. Our results show that active haptic guidance can significantly improve the virtual experience by reducing the number of ``breaks in presence'' and keeping them a safe distance away from physical objects for longer.
    
\end{itemize}

\section{BACKGROUND AND RELATED WORK}
Haptic feedback can be utilized in any mixed reality setting, but in this work we mainly discuss applications of haptics to VR settings, since our implementation was done in VR.
In VR, the user wears a HMD through which they view a 3D, computer-generated VE \cite{jerald2015vr}.
The user's position in the PE is tracked, so that whenever the user updates their position in the PE, the position of the virtual camera updates accordingly to provide an accurate viewing perspective of the VE.
VR is an interactive experience, meaning that the user does not passively observe the virtual content, but instead the environment changes in response to the user's actions and movements.
When the virtual experience feels sufficiently real, the user experiences a sense of presence, which describes the subjective feeling of really being in the environment \cite{slater1997framework}.
Factors that contribute to a user's feelings of presence and immersion in a VE include the HMD refresh rate \cite{barfield1995effect}, the environment realism and visual quality \cite{welch1996effects}, and perceptual stimuli matching \cite{hendrix1996presence, usoh1999walking} (the process of providing users with perceptual information that matches their actions in the VE).
In this paper, we focus on providing haptic stimuli for perceptual stimulus matching to improve the user's experience in VR.

Haptic feedback can be provided in a passive or an active manner.
With passive haptics, objects are placed in the PE such that they align with the locations of objects in the VE, resulting in haptic feedback when the user tries to touch objects in the VE \cite{insko2001passive}.
Conversely, active haptics involves a haptic proxy that dynamically alters its configuration in real time to provide the appropriate haptic force feedback, depending on the user's interactions with the VE.
It is common to use robotic systems to render haptic forces.
For example, Zhang et al. \cite{zhang2023robot} used a robotic arm to provide haptic feedback during object assembly by aligning the arm's end effector with the handheld proxy.
Siu et al. \cite{siu2018shapeshift} used an array of actuated pins to match the contours of virtual objects.
Similarly, other researchers have used tangible, re-configurable electronic peripherals to represent virtual objects \cite{zhao2017robotic,li2022tangiblegrid}.
To recreate the feelings of grasping virtual objects, Kovacs et al.~\cite{kovacs2020haptic} used a wrist-worn device to provide on-demand haptic feedback when users grip virtual objects, while Sinclair et al. \cite{sinclair2019capstancrunch} used a force-resisting, handheld controller to render haptic forces for rigid and compliant objects.
Suzuki et al. \cite{suzuki2020roomshift} used mobile robots to rearrange physical furniture to align with virtual furniture as the user moved through a virtual world.
Robotic systems have also been used to help visually-impaired users understand the virtual content \cite{kunz2018virtual,li2023toucha11y,siu2020virtual,zhao2018enabling}.
Finally, mobile robots have been used to simulate different kinds of virtual terrain that the user walks over \cite{iwata2001gait, iwata2005circulafloor}.

The majority of prior work on active haptics for mixed reality requires the user to initiate interactions with the VE before the haptic forces are rendered.
That is, the haptic forces are triggered by the user's interactions with the VE, so it is the user's actions that dictate when haptic forces are rendered.
In this work, we make the distinction of using active haptics specifically to direct the user and influence their behavior in the VE (in addition to providing a more immersive experience, as all haptics aims to do).
We define this use of haptics as \textit{active haptic guidance}, since it is the haptic forces that direct the user's behaviors, rather than the other way around.
We note that there already exists research on ``haptic guidance,'' which Feygin et al. use to refer to haptic feedback that is used to help people learn motor skills \cite{feygin2002haptic}.
The distinction between our work on active haptic guidance and Feygin et al.'s work is that we use haptic feedback to discreetly influence the user's behavior in an effort to enhance their feelings of presence and level of safety in a mixed reality experience, while Feygin et al. use haptics to teach people motor skills.
\textcolor{black}{Furthermore, haptic guidance by Feygin et al. gives users little autonomy to control their movements during the guidance, whereas our active haptic guidance framework does not take away the user's autonomy to interact with the VE however they wish, which is a crucial component of MR experiences \cite{jerald2015vr}.
Finally, haptic guidance \cite{feygin2002haptic} is designed for real-world use-case, while active haptic guidance is designed specifically for MR experiences.
}

\section{PROBLEM DESCRIPTION}
\label{sec:problem_description}
Here we describe the active haptic guidance problem, as well as  constraints that need to be satisfied to effectively implement active haptic guidance.

\subsection{Definitions}
In virtual reality, the user is located in a PE and VE at the same time. 
Each environment consists of objects (either physical objects or virtual objects represented by textured meshes) and agents (the users and robots).
Note that it is common to refer to virtual humans and animals as agents, but in this work we will consider all components of the VE as generic objects for simplicity, and we use ``agents'' to refer only to humans and robots in the PE.

Let $O = \{o_1, o_2, ..., o_i\}$ be a set of polygonal objects, where each object $o$ is a mesh with vertices in $\mathbb{R}^3$.
Let $U = \{u_1, u_2, ..., u_j\}$ be the set of users in an environment.
Here, $u$ represents the user's state in an environment, and usually describes their position and orientation in said environment.
For example, we can define $u = \{p, \theta\}$, where $p \in \mathbb{R}^2$ represents their position in the 2D plane and $\theta~\in~[0,~2\pi)$ represents their orientation in the environment.
Let $R~=~\{ r_1, r_2, ..., r_k \}$ be the set of robots in an environment, and let $A = \{U \cup R\}$ be the set of all agents.
Each set $O, U, R, \text{and } A$ may be empty.

We define an environment $E$ as a set of obstacles and agents; that is, $E = \{O, A\}$.
To differentiate between the PE and VE, we denote the PE as $E_P = \{O_P, A_P \}$ and the VE as $E_V = \{O_V, A_V \}$.
For each user in VR, they will have a representation in both the PE and VE, so $|U_P| = |U_V| = n$, where $n$ is the number of users.
Since we only consider agents to be users and robots in this work, $|A_V| = n$ (i.e., the only agents in the VE are the users).
In the VE, there are some objects that the user is likely to interact with, which will improve their sense of presence in the environment.
We define this set of ``objects of interest'' $\mathcal{O} \subset O_V$ as the set of virtual objects for which we render haptic forces when the user interacts with them.

With these definitions of the PE and VE, we can now describe the two main conditions that need to be met to provide active haptic guidance to users in MR.
First, the robots in the physical environment need to provide the appropriate haptic feedback to influence the user's configuration.
Second, we need to ensure that the robots that provide haptic feedback are co-located with the virtual objects of interest with which the haptic forces are associated.

\subsection{Influential Haptics Constraint}
\label{subsec:active_guidance_constraint}

The first condition that \textcolor{black}{must} be met is that the {\bf rendered haptic forces should influence the user's behavior such that they update their {\em physical} and {\em virtual} configurations}.
We dub this the \textbf{\textit{influential haptics} (IH) constraint}.
For simplicity, we formalize this constraint using one user, one robot, and one virtual object of interest, but this constraint applies to any group of agents and virtual objects for which we render haptic forces.

Given the user's physical and virtual configurations $u_P$ and $u_V$, a virtual object of interest $o$, and a robot $r$ that provides haptic feedback for $o$, we wish to render a haptic force $\mathbf{F}$ that compels the user to update $u_P$ and $u_V$ to some goal configurations $u_P^*$ and $u_V^*$.
Thus, fulfilling the IH constraint requires completing the following steps:
\begin{enumerate}
    \item Compute the goal configurations $u_P^*$ and $u_V^*$.
    \item Detect or initiate an interaction $I$ between $o$ and $u_V$.
    \item Update the configuration of $r$ to render a haptic force $\mathbf{F}(I, u_V, u_P, u_P^*, u_V^*, r)$ that minimizes an objective function $f(u_V, u_P, u_P^*, u_V^*)$.
\end{enumerate}
In practice, computing $\mathbf{F}(I, u_V, u_P, u_P^*, u_V^*, r)$ depends heavily on the mechanics of $r$ and the objective function $f(u_V, u_P, u_P^*, u_V^*)$.
The objective function is usually a distance function that measures the error between \textcolor{black}{$u_P$ and $u_P^*$, as well as $u_V$ and $u_V^*$ (e.g. Euclidean distance)}, and it depends on the user's configuration space.
By rendering $\mathbf{F}$, the user hopefully updates their configuration in order to move closer to $u_P^*$ and $u_V^*$.

Computing $u_P^*$ and $u_V^*$ is a matter of determining how we want the user to behave, and will vary depending on the MR application context.
In MR, two main reasons to influence the user's behavior are to ensure their safety and to deliver a more immersive experience.
In MR systems, the user tries to navigate through the PE and the VE at the same time, but the PE is partially or fully occluded.
Thus, in order to prevent the user from bumping into physical objects that they cannot see, locomotion interfaces for MR usually display a notification that prompts them to reposition themself to a safer position away from nearby objects.
By using haptics to warn users (either overtly or subtly), we can decrease the likelihood that the user collides with unseen physical obstacles or exits the designated tracking area.
In addition to ensuring user safety, influencing the user's behavior can be useful for improving the user's sense of presence in the VE.
In MR, providing perceptual stimuli that align with the content rendered on the visual display enhances the user's feeling that they are really in the VE that they are seeing.
To this end, haptic feedback can significantly improve the user's sense of presence in the VE \cite{insko2001passive}.
In the case of active haptic guidance, the haptic feedback can be used as an additional narrative element that encourages users to explore a particular area or interact with particular objects in the VE (e.g. pairing visual distractors \cite{peck2009evaluation} with haptic feedback to direct the user's attention).

\subsection{Relative Co-location Constraint}
The second main constraint that should be met when using active haptic guidance is that the {\bf physical robots that render the haptic forces and their associated virtual objects should be co-located {\em relative} to the user}.
That is, the position of the robot and the virtual object should be the same relative to $u_P$ and $u_V$.
This is done to ensure that the user perceives a congruent VE that is augmented by haptic forces, rather than perceiving a VE along with misaligned haptic forces, which may break their sense of presence.
We call this the {\bf \textit{relative co-location} (RC) constraint}.

Given $u_P$, $u_V$, $o$, and $r$ which provides haptic feedback for $o$, we wish to update $r$ such that we minimize the error in the relative positions between $u_V$ and $o$ and $u_P$ and $r$.
Fulfilling the RC constraint requires completing the following steps:
\begin{enumerate}
    \item Compute the configurations of $o$ and $r$ relative to $u_V$ and $u_P$, respectively. Usually, these are just positions $p_o$ and $p_r$ of $o$ and $r$ relative to the user in the respective environment.
    \item Compute a goal configuration $r^*$ for the haptic proxy that minimizes an objective function $f(p_o, p_r)$.
    \item Update the configuration of $r$ to move it towards $r^*$.
\end{enumerate}
In practice, updating the robot's configuration in step \#3 is a motion planning problem where we aim to find a path through the configuration space that brings $r$ close to $r^*$, and it depends on the mechanics of the haptic proxy.

Since MR is an interactive technology, $p_o$ and $p_r$ are constantly changing as the user explores and interacts with the VE.
Thus, evaluating and fulfilling the RC constraint must be done regularly to ensure that any perceptual stimuli mismatch is minimized.
Failure to adequately meet this constraint can degrade the user experience, since it increases the likelihood that the user notices a discrepancy between the visual and haptic stimuli \cite{jansson2004effects,ocampo2019visual}.
Furthermore, knowing how much error between their relative positions the user will tolerate is a subjective measure \cite{azmandian2016haptic, kohli2013redirected}, so it is usually not the case that the robot must reach $r^*$ exactly.
Note that this {\em relative co-location constraint} is not unique to the active haptic guidance problem (unlike \autoref{subsec:active_guidance_constraint}); other work on active haptics for virtual reality also has to deal with the problem of ensuring the co-location of robotic agents and their virtual counterparts \cite{zhang2023robot,vonach2017vrrobot,araujo2016snake,mercado2019entropia}.

\section{PROTOTYPE REALIZATION EXAMPLES}

In this section, we provide details on our prototype implementation of an application of active haptic guidance.
In particular, we implement an \textit{active haptic-driven locomotion} application to provide a safer and more immersive virtual navigation experience for users.
We discuss other potential use-cases for active haptic guidance in the supplementary materials posted on our project page.

\subsection{Natural Walking in Virtual Reality}
In VR, it is common for the PE to be much smaller than the VE.
To enable users to explore large VEs, many different locomotion interfaces such as teleportation, joystick navigation, and walking-in-place have been developed \cite{di2021locomotion}.
Ideally, users explore the VE using natural, everyday walking since it improves their sense of presence \cite{usoh1999walking} and performance in tasks \cite{hodgson2008redirected,peck2011evaluation,ruddle2009benefits}.
One technique that enables natural walking in VR is redirected walking (RDW) \cite{razzaque2001redirected}.

RDW works by slowly rotating the VE around the user's virtual camera while they walk, which causes them to unconsciously adjust their physical trajectory to counteract the VE rotations and remain on their intended path in the VE.
It works because the human perceptual system tends to believe the user's visual stimuli over other stimuli (proprioceptive, vestibular, etc.) when they conflict, as is the case in RDW~\cite{razzaque2005redirected}.
Using RDW, we can steer the user along paths in the PE that direct them away from objects and edges of the tracked space, resulting in a safer and more immersive virtual experience.
To help mask the VE rotations, researchers make use of distractors which grab the user's attention to decrease the likelihood that they attend to the rotations of the VE \cite{chen2017supporting,peck2009evaluation,williams2019estimation}.
In our prototype implementation, we use a virtual dog as a distractor in conjunction with a RDW algorithm known as steer-to-center, which rotates the VE such that the user is steered towards the center of the PE at all times \cite{razzaque2005redirected}.

\subsection{Virtual Experience and Equipment}
For our implementation, a user $u_1$ completed a navigation task in a virtual city and had a virtual dog as a companion (only a single user participated at a time, so $|U_P|~=~|U_V|~=~1$).
Additionally, $u_1$ held a position-tracked leash that was tethered to a differential wheeled robot $r_1$.
The PE was an empty rectangular room with four walls (represented by the boundaries of the VR tracking space).
Thus, $E_P~=~\{ O_P,~A_P \}$, where $A_P~=~\{ u_1,~r_1 \}$.
The virtual dog served as a distractor and was the only object of interest in $E_V$ ($|\mathcal{O}|=1$), meaning that the robot only rendered haptic forces associated with the virtual dog.

Our application was implemented using one HTC VIVE Cosmos VR HMD with two VIVE trackers, and one robot car (ELEGOO UNO Robot Car kit).
We attached one VIVE tracker to the robot to track its location and orientation data, and the other was attached to the leash handle to calculate the distance between $u_1$ and $r_1$. 
The robot was equipped with an HC-06 Bluetooth LE adapter, which connected to the PC to transmit robot movement commands. 
The Unreal 4.22 game engine was used to render the VE.

\subsection{Virtual Companion and Robot Behavior}

Here we describe the behavior of the virtual dog companion and how the robot matches the virtual companion's movements and provides haptic feedback.

\subsubsection{Virtual Dog Companion Behavior}
The virtual dog has two main behavior states: \textit{following} and \textit{distracting}.
When the user walks around and is not at risk of leaving the tracking space, the dog is in \textit{follow} mode.
In this mode, the dog walks slightly ahead of the user as they walk, and remains in one spot when the user stands still.

When the user reaches a boundary of the tracked space, the VR system initiates what is called a \textit{reset}, wherein the user reorients themself such that they face towards the inside of the tracking space in the PE.
To ensure that their orientation in the VE is not altered, the VR system applies redirection that effectively cancels out their physical rotation in the virtual space.
When a reset is initiated, the virtual dog enters \textit{distract} mode.
In \textit{distract} mode, we compute a goal position in the VE for the dog to move towards.
The idea behind \textit{distract} mode is that the user is likely to pay attention to the virtual dog as it runs to a goal position, which allows the system to apply stronger redirection (away from the obstacles in the PE) without interfering with the user's experience \cite{peck2009evaluation}.

\begin{figure*}[!htb]
  \centering
  \vspace*{-0.5em}
  \includegraphics[width=0.99\textwidth]{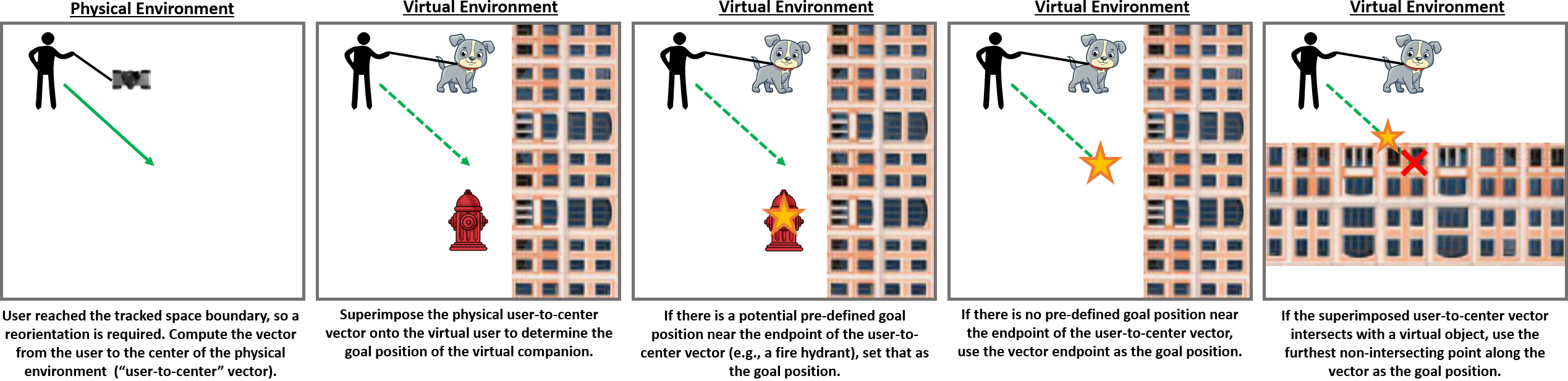}
  \vspace*{-0.5em}
  \caption{Our method of automatically choosing a suitable virtual goal position for the virtual companion. When the user gets close to a boundary of the physical space, they need to be reoriented away from the boundary before they continue walking. In order to pick a goal destination for the virtual companion and robotic haptic proxy, we cast a ray from the physical user to the center of the tracked space and then superimpose this vector onto the user's virtual position. If the endpoint of this vector is near a pre-defined potential goal position, that is chosen as the current goal position. Otherwise, we choose the furthest point along the vector that does not intersect with any objects in the virtual environment.}
  \label{fig:point}
  \vspace*{-1em}
\end{figure*}

During a reset, the goal position is selected by first computing the vector from the user towards the center of the physical space.
The goal position is then determined to be either the endpoint of this vector in the VE, or a virtual object near the vector's endpoint that was labeled as a potential goal position during development.
Potential goal positions are virtual objects that a dog would be likely to interact with, such as a fire hydrant or a lamp post.
If the vector intersects with a virtual object (e.g. a virtual building) and there are no potential goal objects nearby, the goal position is simply the point furthest along the vector that does not intersect with any objects.
See \autoref{fig:point} for a visualization of this process.

\subsubsection{Robot Haptic Proxy Behavior}
\label{subsubsec:robot_behavior}
The physical robot's main purpose is to provide haptic feedback to make the user's virtual experience feel more immersive and to encourage the user to walk away from nearby objects or tracking space boundaries in the PE.
In both \textit{follow} and \textit{distract} mode, the physical robot needs to update its position such that it is aligned with the position of the virtual dog, relative to the user in either environment.
Checking if a position update is necessary is easily achieved by computing the vector from the virtual user to the virtual dog and comparing it to the vector from the user's HMD and the robot.

To compute the trajectory that the robot will follow, we compute a circular arc path based on the robot's position, forward direction, and destination position (determined by the relative position of the virtual dog and user).
The ideal path for a differential drive robot is a circular arc since it only requires one set of wheel velocities \cite{chitsaz2009minimum}. 
The wheel velocities are computed with the ratio $\frac{2rd}{2r-d}$, where $r$ is the arc radius and $d$ is the distance between the robot wheels.
Note that we do not use typical PID-based drift correction due to possible unexpected complications that may arise from the tethering to the user \cite{aaarzen1999simple,rivera1986internal,skogestad2003simple}.

\subsection{Maintaining Active Haptic Guidance Constraints}
This section describes how our active haptic-drive locomotion application satisfies the IH and RC constraints.

\subsubsection{Directing Users With Haptic Feedback}
Since the virtual object of interest is a dog, the user is attached to the robot by an elastic tether that resembles a leash.
When the robot moves away from the user in the PE, it simulates the sensation of a dog tugging on its leash, thereby improving the realism of the virtual experience.
Additionally, this tugging encourages the user to follow the robot rather than ``fight'' it, allowing us to further influence the user's movement patterns in the PE and VE.
By triggering the robot to move away from the user and towards the center of the PE when they get too close to the tracking space boundaries, the tugging force on the leash encourages the user to turn and walk towards the robot and away from the tracked space boundaries.

\subsubsection{Maintaining Co-location}
Normally, maintaining relative co-location between a haptic proxy and a virtual object is a matter of updating the position of the haptic proxy when the virtual object's position changes.
In our implementation, we update the position of the robot to match the movements of the virtual dog.
However, our implementation requires additional work to maintain co-location due to a problem that we call the \textit{haptic proxy distortion} (HPD) problem.

\begin{figure}[!htb]
  \centering
  \vspace*{0.5em}
  \includegraphics[width=0.95\linewidth]{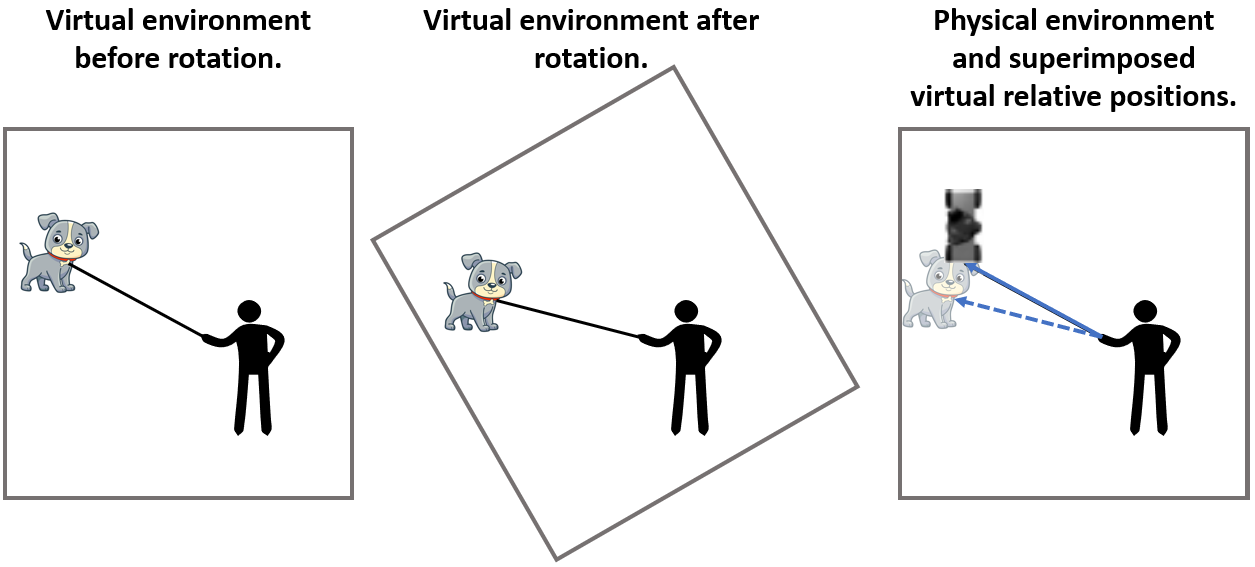}
  \vspace*{-1em}
  \caption{A visualization of the haptic proxy distortion problem. \textit{Left:} Initially, the virtual user and virtual companion have a particular relative position. \textit{Middle:} After rotating the virtual environment around the virtual user, the relative position of the companion changes since the companion is rotated along with the rest of the environment. \textit{Right:} In the physical space, the haptic proxy has not been updated, so its position coincides with the virtual companion's relative position \textit{before} rotation (opaque robot and vector). The new relative position of the virtual companion, which the haptic proxy needs to match, is shown as the translucent dog and dashed-line vector.}
  \label{fig:hpd_problem}
\end{figure}

In our implementation, we \textcolor{black}{build on the RDW locomotion interface to enable natural walking in VR}.
RDW works by rotating the entire VE around the virtual camera that represents the user's viewpoint in VR.
Consequently, the virtual dog companion may change its position relative to the virtual user \textit{without the dog actually moving to a new destination in the VE} (see \autoref{fig:hpd_problem}).
As we apply redirection, the relative position of the virtual dog changes constantly while the relative position of the physical robot does not.
To resolve this discrepancy in relative position, we update the robot's goal destination in the PE on every frame to minimize the difference in relative position.
The user will perceive this as the haptic proxy ``sliding'' across the floor around them, which might result in unsmooth motion that may detract from the user experience.
In practice, this did not seem to be a major problem for users, but we acknowledge that there may be better solutions to the HPD problem, and leave that for future work.
This HPD problem adds onto the errors in relative co-location between the haptic proxy and the virtual companion, which makes it harder to satisfy the RC constraint.

\section{EXPERIMENTS \& RESULTS}
\label{sec:experiment}

\subsection{Experiment Design and Procedure}
To evaluate the effectiveness of our implementation of active haptic-driven locomotion prototype, we conducted a user study where participants completed a navigation task.
The goal of our user study was to evaluate how effective the haptic guidance was at improving users' sense of presence in the VE and keeping users away from the boundaries of the VR system's tracked space.
We used a between participants design, where one group of participants completed a navigation task with active haptic guidance enabled, and the other group completed the same task without any haptic guidance.
\textcolor{black}{
We chose a between-subjects design to avoid potential learning effects in the navigation task and to make the experiment duration more manageable.}
A total of 20 participants (13 male, 5 female, 2 participants did not report) completed our experiment (age $\mu = 24.59$, $\sigma = 2.37$).
All participants were able to walk \textcolor{black}{unassisted when not using VR}.

During the experiment, participants were placed in a virtual city environment with several streets and blocks (see \autoref{fig:teaser} and our video).
Participants started the task at one intersection in the city, and their task was to reach a green question mark in the environment that indicated their destination, which was one block away from the their starting position.
During the experiment, we recorded how many times users reached the bounds of the PE and the time taken to complete the task.
Once participants finished the task, they completed a questionnaire with questions on a 7-point Likert scale that measured their sense of presence in the VE (7 = high presence, 1 = low presence).
Finally, the experiment was ended with open-ended questions where participants could provide additional comments.
The study lasted about 15 minutes in total.

\subsection{Results and Discussion}
The metrics we used to measure the effectiveness of our active haptic-driven locomotion interface were the number of breaks in presence (BiPs), the completion rate and time taken to complete the task, and participants' subjective feelings of presence in the VE.
A BiP is incurred when the user reaches the boundaries of the tracking space and they are forced to reorient away from the boundary before continuing to walk.

\textcolor{black}{We conducted Mann–Whitney U tests to compare the differences in performance metrics between the two conditions (see \autoref{tab:user_study_results}).}
The presence of our active haptic guidance companion resulted in significantly fewer BiPs, lower completion times and higher completion rates, and slightly higher (and above-average) presence levels.
Meanwhile, participants who completed the navigation task without any haptic guidance incurred a large number of BiPs, did not finish the task in time, and reported below-average levels of presence.
\textcolor{black}{Note that although RDW is an established method for locomotion in VR, its usefulness is limited if the PE is too small, since there will not be enough space for the user's physical and virtual paths to significantly deviate such that they can avoid physical obstacles \cite{suma2012impossible}.}
These results support the notion that active haptic guidance can be used to help keep users safe and feel more immersed in mixed reality experiences.

\textcolor{black}{The results in \autoref{tab:user_study_results} mainly evaluate the IH constraint and do not provide much insight into the RC constraint. 
In practice, it can be difficult to accurately measure how well the RC constraint is met, since it depends on the user's subjective tolerance of the mismatch between haptic and visual stimuli as well as the particular trajectory the user travels on.
The bluetooth adapter that received movement commands from the computer had a baud rate of 9600, so it was capable of updating its position at the frame rate of our VR application (90 frames per second).
During the informal questioning after the experiment, some participants reported that they found the dog to be well-behaved and they enjoyed the haptic feedback it provided, which suggests that the RC constraint was adequately met.
}

\setlength{\tabcolsep}{4.5pt}
\begin{table}[t]
    \centering
    
    \caption{Performance results from our user study.} 
    
    \begin{tabular}{l|llllllll}
         & \multicolumn{2}{c}{\textbf{BiPs}} & \multicolumn{2}{c}{\textbf{Time} (s)} & \multicolumn{2}{c}{\textbf{Presence}} & \multicolumn{2}{c}{\textbf{Completed}} \\ \midrule \midrule
        \textbf{Haptics} & $\mu$ & $\sigma$ & $\mu$ & $\sigma$ & $\mu$ & $\sigma$ & \multicolumn{2}{c}{Total \#} \\ \midrule
        With & {\bf 0.90} & 0.74 & {\bf 195.20} & 22.25 & {\bf 4.63} & 1.77 & \multicolumn{2}{c}{\bf 10} \\ 
        Without & 18.90 & 5.17 & 309.40 & 65.14 & 3.57 & 1.64 & \multicolumn{2}{c}{1} \\ 
        \midrule
        \multicolumn{2}{l}{\textbf{Significance}} & \multicolumn{1}{c}{} & \multicolumn{2}{c}{} & \multicolumn{2}{c}{} & \multicolumn{2}{c}{} \\
        \midrule
        $p$-value & \multicolumn{2}{c}{$p<.0001$} & \multicolumn{2}{c}{$p=0.0008$} & \multicolumn{2}{c}{$p<.0001$} & \multicolumn{2}{c}{$p<.0001$} \\
        Statistic & \multicolumn{2}{c}{$U=0.0$} & \multicolumn{2}{c}{$U=10.0$} & \multicolumn{2}{c}{$U=4010.5$} & \multicolumn{2}{c}{$5.0$} \\
        \midrule \midrule
    \end{tabular}
    
    \label{tab:user_study_results}
\end{table}

\section{CONCLUSIONS \& FUTURE WORK}

In this work, we presented the \textit{active haptic guidance} problem for MR, which describes the use of one or more robots to provide haptic feedback to users in order to create a richer virtual experience for them, while \textit{also} influencing their behavior to improve users' safety and immersion in the VE.
We implemented active haptic guidance in a VR locomotion application that enabled the user to explore a large VE while located in a much smaller PE.
By combining active haptic guidance and RDW, we increased the effective area of the PE while also decreasing the likelihood that the user exits the VR system's tracked area.
The concept of active haptic guidance is general and can be applied MR applications other than locomotion; we discuss other potential use cases for active haptic guidance in the supplementary materials.
One limitation of our work is the haptic proxy distortion problem, in which the haptic proxy and the associated virtual object can become misaligned due to mismatches between the user's physical and virtual configurations.
Solving this problem requires continuously updating the position of the haptic proxy, and our proposed solution in this work is likely not the most optimal solution.
Additionally, our system uses an estimation of drift to update the haptic proxy position, instead of a more accurate method like PID-based drift correction.
Future work in this area should study the use of more realistic companion behaviors, and should explore how active haptic guidance can be applied to other immersive experiences like social MR settings with other users.

{\small
\bibliographystyle{IEEEtranS}
\bibliography{references}
}



\newpage
\newpage
\clearpage

\section*{APPENDIX}

\subsection{Additional Experiment Design and Procedure Details}
\textcolor{black}{For participants who completed the task without haptic guidance, they were instructed to stop walking and turn around in place until an indicator of the boundary's proximity disappeared, at which point they were free to continue walking towards the goal. 
During this reorientation procedure, redirection was applied to ensure that their virtual heading direction was the same before and after the reorientation (to avoid interfering with their navigation process).}
The navigation task had a time limit of 5 minutes and 30 seconds, after which the experiment ended regardless of if the participant reached the goal destination.
Participants were unaware of this time limit so that they did not rush to complete the task.

Before participants put on the HMD, we debriefed them on the experiment procedures and had them complete a pre-study Simulator Sickness Questionnaire (SSQ) \cite{kennedy1993simulator}.
Next, the user put on the HMD and completed the task in the VE.
The VE was a city environment with several streets and blocks, and was populated with common objects such as bus stops, stores, park squares, and virtual humans that roamed around the environment (see \autoref{fig:teaser} and our video).
To mask any potentially distracting noises from the robot as it moves, participants wore headphones and background music was played for the duration of the experiment task.
Once participants finished the task, they completed another SSQ survey in addition to the sense of presence questionnaire and answered some open-ended questions.
The study design was approved by our university's Institutional Review Board.

\subsection{Additional Applications of Active Haptic Guidance}

Here we discuss other potential applications of active haptic guidance for immersive applications:
\begin{itemize}
    \item \textbf{Wood Carving Application:} In wood carving, the grain of the wood will impact the direction in which the artist carves the wood. That is, sometimes the artist will carve ``with the grain'' and sometimes will carve ``against the grain.'' Using active haptics, one could accurately render the different resistance forces that arise from carving with or against the grain of a virtual wooden block, which will in turn influence the way in which the user carves their virtual wooden sculpture. In addition to providing a more realistic experience, this could be used to guide the user to create a more appealing final sculpture (e.g. by altering the direction of the grain to subtly change their hand movements, which will change the shape of the final carved surface).
    
    \item \textbf{Immersive Cooperative Application:} A major appeals of mixed reality experiences is the ability to connect with other users in shared virtual experiences. Important to these shared experiences is the ability to touch the other person, which can provide a greater sense of companionship and connection between users. Haptic forces can be used to encourage users to interact with or follow other users who are also present in their virtual experience, which may improve the users' sense of presence in the experience due to the enhanced realism.
    
    \item \textbf{Virtual Cooking Training Application:} Given a seated VR experience where the user is practicing their cooking skills in a virtual environment, a mobile, tabletop robot can provide haptic feedback that represents feedback provided by cooking utensils.
    For example, when spreading brownie batter in a baking pan, the user will feel haptic forces when the virtual spreading utensil gets too close to the edges of the virtual baking pan.
    These forces could be rendered using a mobile robot with a flat surface that serves as a wall that the user's physical hand will bump into.
\end{itemize}




\end{document}